\documentclass{article}
\def\false{|\!\!\not\models}
\def\bel#1#2{{\cal B}_{#1}(#2)}
\def\des#1#2{{\cal W}_{#1}(#2)}
\def\int#1#2{{\cal I}_{#1}(#2)}
\def\know#1#2{{\cal K}_{#1}(#2)}
\def\action#1#2{{\bf do}_{#1}(#2)}
\def\constative#1{\mbox{constative}(#1)}
\def\interrogative#1{\mbox{interrogative}(#1)}
\def\imperative#1{\mbox{imperative}(#1)}
\def\inform#1#2{\mbox{\sf inform}_{#1}(#2)}
\def\query#1#2{\mbox{\sf query}_{#1}(#2)}
\def\cancel#1#2{\mbox{\sf cancel}_{#1}(#2)}

\def\newand#1{\mbox{\bf newAND}(#1)}
\def\newor#1{\mbox{\bf newOR}(#1)}
\def\newsubtree#1{\mbox{\bf newSubTree}(#1)}
\def\newnode#1{\mbox{\bf newNode}(#1)}
\def\changeroot#1{\mbox{\bf changeRoot}(#1)}
\def\system#1#2{\par \noindent \leavevmode \hbox to.75cm{#1\hss}%
\bgroup \hangafter=1
\hangindent=1cm {\bf System:} #2\par \egroup}
\def\user#1#2{\par \noindent \bgroup \hangafter=1
\hangindent=1cm \leavevmode \hbox to.75cm{#1\hss}%
 {\bf User:} #2 \par \egroup}
\def\subst#1\with#2\endsubst{\ifvmode \par \leavevmode \fi
#2
\ignorespaces
\setbox\revisions=\vbox{\unvbox\revisions \vskip\baselineskip \noindent
{\bf Auf Seite \thepage} \vskip.25\baselineskip% #1
\vskip.25\baselineskip \noindent {\bf ersetzt
durch}\vskip.25\baselineskip #2\vskip.25\baselineskip}}
\def\newtext#1\endsubst{\ifvmode \par \leavevmode \fi
#1
\ignorespaces
\setbox\revisions=\vbox{\unvbox\revisions \vskip\baselineskip \noindent
{\bf Auf Seite \thepage\ eingef\"ugt} \vskip.25\baselineskip
#1\vskip.25\baselineskip}}

\usepackage{a4wide}
\newcounter{treecount}
\newcounter{branchcount}
\newsavebox{\parentbox}
\newsavebox{\treebox}
\newsavebox{\treeboxone}
\newsavebox{\treeboxtwo}
\newsavebox{\treeboxthree}
\newsavebox{\treeboxfour}
\newsavebox{\treeboxfive}
\newsavebox{\treeboxsix}
\newsavebox{\treeboxseven}
\newsavebox{\treeboxeight}
\newsavebox{\treeboxnine}
\newsavebox{\treeboxten}
\newsavebox{\treeboxeleven}
\newsavebox{\treeboxtwelve}
\newsavebox{\treeboxthirteen}
\newsavebox{\treeboxfourteen}
\newsavebox{\treeboxfifteen}
\newsavebox{\treeboxsixteen}
\newsavebox{\treeboxseventeen}
\newsavebox{\treeboxeighteen}
\newsavebox{\treeboxnineteen}
\newsavebox{\treeboxtwenty}
\newlength{\treeoffsetone}
\newlength{\treeoffsettwo}
\newlength{\treeoffsetthree}
\newlength{\treeoffsetfour}
\newlength{\treeoffsetfive}
\newlength{\treeoffsetsix}
\newlength{\treeoffsetseven}
\newlength{\treeoffseteight}
\newlength{\treeoffsetnine}
\newlength{\treeoffsetten}
\newlength{\treeoffseteleven}
\newlength{\treeoffsettwelve}
\newlength{\treeoffsetthirteen}
\newlength{\treeoffsetfourteen}
\newlength{\treeoffsetfifteen}
\newlength{\treeoffsetsixteen}
\newlength{\treeoffsetseventeen}
\newlength{\treeoffseteighteen}
\newlength{\treeoffsetnineteen}
\newlength{\treeoffsettwenty}

\newlength{\treeshiftone}
\newlength{\treeshifttwo}
\newlength{\treeshiftthree}
\newlength{\treeshiftfour}
\newlength{\treeshiftfive}
\newlength{\treeshiftsix}
\newlength{\treeshiftseven}
\newlength{\treeshifteight}
\newlength{\treeshiftnine}
\newlength{\treeshiftten}
\newlength{\treeshifteleven}
\newlength{\treeshifttwelve}
\newlength{\treeshiftthirteen}
\newlength{\treeshiftfourteen}
\newlength{\treeshiftfifteen}
\newlength{\treeshiftsixteen}
\newlength{\treeshiftseventeen}
\newlength{\treeshifteighteen}
\newlength{\treeshiftnineteen}
\newlength{\treeshifttwenty}
\newlength{\treewidthone}
\newlength{\treewidthtwo}
\newlength{\treewidththree}
\newlength{\treewidthfour}
\newlength{\treewidthfive}
\newlength{\treewidthsix}
\newlength{\treewidthseven}
\newlength{\treewidtheight}
\newlength{\treewidthnine}
\newlength{\treewidthten}
\newlength{\treewidtheleven}
\newlength{\treewidthtwelve}
\newlength{\treewidththirteen}
\newlength{\treewidthfourteen}
\newlength{\treewidthfifteen}
\newlength{\treewidthsixteen}
\newlength{\treewidthseventeen}
\newlength{\treewidtheighteen}
\newlength{\treewidthnineteen}
\newlength{\treewidthtwenty}
\newlength{\daughteroffsetone}
\newlength{\daughteroffsettwo}
\newlength{\daughteroffsetthree}
\newlength{\daughteroffsetfour}
\newlength{\branchwidthone}
\newlength{\branchwidthtwo}
\newlength{\branchwidththree}
\newlength{\branchwidthfour}
\newlength{\parentoffset}
\newlength{\treeoffset}
\newlength{\daughteroffset}
\newlength{\branchwidth}
\newlength{\parentwidth}
\newlength{\treewidth}
\newcommand{\ontop}[1]{\begin{tabular}{c}#1\end{tabular}}
\newcommand{\poptree}{%
\ifnum\value{treecount}=0\typeout{QobiTeX warning---Tree stack underflow}\fi%
\addtocounter{treecount}{-1}%
\setlength{\treeoffsettwo}{\treeoffsetthree}%
\setlength{\treeoffsetthree}{\treeoffsetfour}%
\setlength{\treeoffsetfour}{\treeoffsetfive}%
\setlength{\treeoffsetfive}{\treeoffsetsix}%
\setlength{\treeoffsetsix}{\treeoffsetseven}%
\setlength{\treeoffsetseven}{\treeoffseteight}%
\setlength{\treeoffseteight}{\treeoffsetnine}%
\setlength{\treeoffsetnine}{\treeoffsetten}%
\setlength{\treeoffsetten}{\treeoffseteleven}%
\setlength{\treeoffseteleven}{\treeoffsettwelve}%
\setlength{\treeoffsettwelve}{\treeoffsetthirteen}%
\setlength{\treeoffsetthirteen}{\treeoffsetfourteen}%
\setlength{\treeoffsetfourteen}{\treeoffsetfifteen}%
\setlength{\treeoffsetfifteen}{\treeoffsetsixteen}%
\setlength{\treeoffsetsixteen}{\treeoffsetseventeen}%
\setlength{\treeoffsetseventeen}{\treeoffseteighteen}%
\setlength{\treeoffseteighteen}{\treeoffsetnineteen}%
\setlength{\treeoffsetnineteen}{\treeoffsettwenty}%
\setlength{\treeshifttwo}{\treeshiftthree}%
\setlength{\treeshiftthree}{\treeshiftfour}%
\setlength{\treeshiftfour}{\treeshiftfive}%
\setlength{\treeshiftfive}{\treeshiftsix}%
\setlength{\treeshiftsix}{\treeshiftseven}%
\setlength{\treeshiftseven}{\treeshifteight}%
\setlength{\treeshifteight}{\treeshiftnine}%
\setlength{\treeshiftnine}{\treeshiftten}%
\setlength{\treeshiftten}{\treeshifteleven}%
\setlength{\treeshifteleven}{\treeshifttwelve}%
\setlength{\treeshifttwelve}{\treeshiftthirteen}%
\setlength{\treeshiftthirteen}{\treeshiftfourteen}%
\setlength{\treeshiftfourteen}{\treeshiftfifteen}%
\setlength{\treeshiftfifteen}{\treeshiftsixteen}%
\setlength{\treeshiftsixteen}{\treeshiftseventeen}%
\setlength{\treeshiftseventeen}{\treeshifteighteen}%
\setlength{\treeshifteighteen}{\treeshiftnineteen}%
\setlength{\treeshiftnineteen}{\treeshifttwenty}%
\setlength{\treewidthtwo}{\treewidththree}%
\setlength{\treewidththree}{\treewidthfour}%
\setlength{\treewidthfour}{\treewidthfive}%
\setlength{\treewidthfive}{\treewidthsix}%
\setlength{\treewidthsix}{\treewidthseven}%
\setlength{\treewidthseven}{\treewidtheight}%
\setlength{\treewidtheight}{\treewidthnine}%
\setlength{\treewidthnine}{\treewidthten}%
\setlength{\treewidthten}{\treewidtheleven}%
\setlength{\treewidtheleven}{\treewidthtwelve}%
\setlength{\treewidthtwelve}{\treewidththirteen}%
\setlength{\treewidththirteen}{\treewidthfourteen}%
\setlength{\treewidthfourteen}{\treewidthfifteen}%
\setlength{\treewidthfifteen}{\treewidthsixteen}%
\setlength{\treewidthsixteen}{\treewidthseventeen}%
\setlength{\treewidthseventeen}{\treewidtheighteen}%
\setlength{\treewidtheighteen}{\treewidthnineteen}%
\setlength{\treewidthnineteen}{\treewidthtwenty}%
\sbox{\treeboxtwo}{\usebox{\treeboxthree}}%
\sbox{\treeboxthree}{\usebox{\treeboxfour}}%
\sbox{\treeboxfour}{\usebox{\treeboxfive}}%
\sbox{\treeboxfive}{\usebox{\treeboxsix}}%
\sbox{\treeboxsix}{\usebox{\treeboxseven}}%
\sbox{\treeboxseven}{\usebox{\treeboxeight}}%
\sbox{\treeboxeight}{\usebox{\treeboxnine}}%
\sbox{\treeboxnine}{\usebox{\treeboxten}}%
\sbox{\treeboxten}{\usebox{\treeboxeleven}}%
\sbox{\treeboxeleven}{\usebox{\treeboxtwelve}}%
\sbox{\treeboxtwelve}{\usebox{\treeboxthirteen}}%
\sbox{\treeboxthirteen}{\usebox{\treeboxfourteen}}%
\sbox{\treeboxfourteen}{\usebox{\treeboxfifteen}}%
\sbox{\treeboxfifteen}{\usebox{\treeboxsixteen}}%
\sbox{\treeboxsixteen}{\usebox{\treeboxseventeen}}%
\sbox{\treeboxseventeen}{\usebox{\treeboxeighteen}}%
\sbox{\treeboxeighteen}{\usebox{\treeboxnineteen}}%
\sbox{\treeboxnineteen}{\usebox{\treeboxtwenty}}}
\newcommand{\leaf}[1]{%
\ifnum\value{treecount}=20\typeout{QobiTeX warning---Tree stack overflow}\fi%
\addtocounter{treecount}{1}%
\sbox{\treeboxtwenty}{\usebox{\treeboxnineteen}}%
\sbox{\treeboxnineteen}{\usebox{\treeboxeighteen}}%
\sbox{\treeboxeighteen}{\usebox{\treeboxseventeen}}%
\sbox{\treeboxseventeen}{\usebox{\treeboxsixteen}}%
\sbox{\treeboxsixteen}{\usebox{\treeboxfifteen}}%
\sbox{\treeboxfifteen}{\usebox{\treeboxfourteen}}%
\sbox{\treeboxfourteen}{\usebox{\treeboxthirteen}}%
\sbox{\treeboxthirteen}{\usebox{\treeboxtwelve}}%
\sbox{\treeboxtwelve}{\usebox{\treeboxeleven}}%
\sbox{\treeboxeleven}{\usebox{\treeboxten}}%
\sbox{\treeboxten}{\usebox{\treeboxnine}}%
\sbox{\treeboxnine}{\usebox{\treeboxeight}}%
\sbox{\treeboxeight}{\usebox{\treeboxseven}}%
\sbox{\treeboxseven}{\usebox{\treeboxsix}}%
\sbox{\treeboxsix}{\usebox{\treeboxfive}}%
\sbox{\treeboxfive}{\usebox{\treeboxfour}}%
\sbox{\treeboxfour}{\usebox{\treeboxthree}}%
\sbox{\treeboxthree}{\usebox{\treeboxtwo}}%
\sbox{\treeboxtwo}{\usebox{\treeboxone}}%
\sbox{\treeboxone}{\ontop{#1}}%
\sbox{\treeboxone}{\raisebox{-\ht\treeboxone}{\usebox{\treeboxone}}}%
\setlength{\treeoffsettwenty}{\treeoffsetnineteen}%
\setlength{\treeoffsetnineteen}{\treeoffseteighteen}%
\setlength{\treeoffseteighteen}{\treeoffsetseventeen}%
\setlength{\treeoffsetseventeen}{\treeoffsetsixteen}%
\setlength{\treeoffsetsixteen}{\treeoffsetfifteen}%
\setlength{\treeoffsetfifteen}{\treeoffsetfourteen}%
\setlength{\treeoffsetfourteen}{\treeoffsetthirteen}%
\setlength{\treeoffsetthirteen}{\treeoffsettwelve}%
\setlength{\treeoffsettwelve}{\treeoffseteleven}%
\setlength{\treeoffseteleven}{\treeoffsetten}%
\setlength{\treeoffsetten}{\treeoffsetnine}%
\setlength{\treeoffsetnine}{\treeoffseteight}%
\setlength{\treeoffseteight}{\treeoffsetseven}%
\setlength{\treeoffsetseven}{\treeoffsetsix}%
\setlength{\treeoffsetsix}{\treeoffsetfive}%
\setlength{\treeoffsetfive}{\treeoffsetfour}%
\setlength{\treeoffsetfour}{\treeoffsetthree}%
\setlength{\treeoffsetthree}{\treeoffsettwo}%
\setlength{\treeoffsettwo}{\treeoffsetone}%
\setlength{\treeoffsetone}{0.5\wd\treeboxone}%
\setlength{\treeshifttwenty}{\treeshiftnineteen}%
\setlength{\treeshiftnineteen}{\treeshifteighteen}%
\setlength{\treeshifteighteen}{\treeshiftseventeen}%
\setlength{\treeshiftseventeen}{\treeshiftsixteen}%
\setlength{\treeshiftsixteen}{\treeshiftfifteen}%
\setlength{\treeshiftfifteen}{\treeshiftfourteen}%
\setlength{\treeshiftfourteen}{\treeshiftthirteen}%
\setlength{\treeshiftthirteen}{\treeshifttwelve}%
\setlength{\treeshifttwelve}{\treeshifteleven}%
\setlength{\treeshifteleven}{\treeshiftten}%
\setlength{\treeshiftten}{\treeshiftnine}%
\setlength{\treeshiftnine}{\treeshifteight}%
\setlength{\treeshifteight}{\treeshiftseven}%
\setlength{\treeshiftseven}{\treeshiftsix}%
\setlength{\treeshiftsix}{\treeshiftfive}%
\setlength{\treeshiftfive}{\treeshiftfour}%
\setlength{\treeshiftfour}{\treeshiftthree}%
\setlength{\treeshiftthree}{\treeshifttwo}%
\setlength{\treeshifttwo}{\treeshiftone}%
\setlength{\treeshiftone}{0pt}%
\setlength{\treewidthtwenty}{\treewidthnineteen}%
\setlength{\treewidthnineteen}{\treewidtheighteen}%
\setlength{\treewidtheighteen}{\treewidthseventeen}%
\setlength{\treewidthseventeen}{\treewidthsixteen}%
\setlength{\treewidthsixteen}{\treewidthfifteen}%
\setlength{\treewidthfifteen}{\treewidthfourteen}%
\setlength{\treewidthfourteen}{\treewidththirteen}%
\setlength{\treewidththirteen}{\treewidthtwelve}%
\setlength{\treewidthtwelve}{\treewidtheleven}%
\setlength{\treewidtheleven}{\treewidthten}%
\setlength{\treewidthten}{\treewidthnine}%
\setlength{\treewidthnine}{\treewidtheight}%
\setlength{\treewidtheight}{\treewidthseven}%
\setlength{\treewidthseven}{\treewidthsix}%
\setlength{\treewidthsix}{\treewidthfive}%
\setlength{\treewidthfive}{\treewidthfour}%
\setlength{\treewidthfour}{\treewidththree}%
\setlength{\treewidththree}{\treewidthtwo}%
\setlength{\treewidthtwo}{\treewidthone}%
\setlength{\treewidthone}{\wd\treeboxone}}
\newcommand{\branch}[2]{%
\setcounter{branchcount}{#1}%
\ifnum\value{branchcount}=1\sbox{\parentbox}{\ontop{#2}}%
\setlength{\parentoffset}{\treeoffsetone}%
\addtolength{\parentoffset}{-0.5\wd\parentbox}%
\setlength{\daughteroffset}{0in}%
\ifdim\parentoffset<0in%
\setlength{\daughteroffset}{-\parentoffset}%
\setlength{\parentoffset}{0in}\fi%
\setlength{\parentwidth}{\parentoffset}%
\addtolength{\parentwidth}{\wd\parentbox}%
\setlength{\treeoffset}{\daughteroffset}%
\addtolength{\treeoffset}{\treeoffsetone}%
\setlength{\treewidth}{\wd\treeboxone}%
\addtolength{\treewidth}{\daughteroffset}%
\ifdim\treewidth<\parentwidth\setlength{\treewidth}{\parentwidth}\fi%
\sbox{\treebox}{\begin{minipage}{\treewidth}%
\begin{flushleft}%
\hspace*{\parentoffset}\usebox{\parentbox}\\
{\setlength{\unitlength}{2ex}%
\hspace*{\treeoffset}\begin{picture}(0,1)%
\put(0,0){\line(0,1){1}}%
\end{picture}}\\
\vspace{-\baselineskip}
\hspace*{\daughteroffset}%
\raisebox{-\ht\treeboxone}{\usebox{\treeboxone}}%
\end{flushleft}%
\end{minipage}}%
\setlength{\treeoffsetone}{\parentoffset}%
\addtolength{\treeoffsetone}{0.5\wd\parentbox}%
\setlength{\treeshiftone}{0pt}%
\setlength{\treewidthone}{\treewidth}%
\sbox{\treeboxone}{\usebox{\treebox}}%
\else\ifnum\value{branchcount}=2\sbox{\parentbox}{\ontop{#2}}%
\setlength{\branchwidthone}{\treewidthtwo}%
\addtolength{\branchwidthone}{\treeoffsetone}%
\addtolength{\branchwidthone}{-\treeshiftone}%
\addtolength{\branchwidthone}{-\treeoffsettwo}%
\setlength{\branchwidth}{\branchwidthone}%
\setlength{\daughteroffsetone}{\branchwidth}%
\addtolength{\daughteroffsetone}{-\branchwidthone}%
\addtolength{\daughteroffsetone}{-\treeshiftone}%
\setlength{\parentoffset}{-0.5\wd\parentbox}%
\addtolength{\parentoffset}{\treeoffsettwo}%
\addtolength{\parentoffset}{0.5\branchwidth}%
\setlength{\daughteroffset}{0in}%
\ifdim\parentoffset<0in%
\setlength{\daughteroffset}{-\parentoffset}%
\setlength{\parentoffset}{0in}\fi%
\setlength{\parentwidth}{\parentoffset}%
\addtolength{\parentwidth}{\wd\parentbox}%
\setlength{\treeoffset}{\daughteroffset}%
\addtolength{\treeoffset}{\treeoffsettwo}%
\setlength{\treewidth}{\wd\treeboxone}%
\addtolength{\treewidth}{\daughteroffsetone}%
\addtolength{\treewidth}{\treewidthtwo}%
\addtolength{\treewidth}{\daughteroffset}%
\ifdim\treewidth<\parentwidth\setlength{\treewidth}{\parentwidth}\fi%
\sbox{\treebox}{\begin{minipage}{\treewidth}%
\begin{flushleft}%
\hspace*{\parentoffset}\usebox{\parentbox}\\
{\setlength{\unitlength}{0.5\branchwidth}%
\hspace*{\treeoffset}\begin{picture}(2,0.5)%
\put(0,0){\line(2,1){1}}%
\put(2,0){\line(-2,1){1}}%
\end{picture}}\\
\vspace{-\baselineskip}
\hspace*{\daughteroffset}%
\makebox[\treewidthtwo][l]%
{\raisebox{-\ht\treeboxtwo}{\usebox{\treeboxtwo}}}%
\hspace*{\daughteroffsetone}%
\raisebox{-\ht\treeboxone}{\usebox{\treeboxone}}%
\end{flushleft}%
\end{minipage}}%
\setlength{\treeoffsetone}{\parentoffset}%
\addtolength{\treeoffsetone}{0.5\wd\parentbox}%
\setlength{\treeshiftone}{0pt}%
\setlength{\treewidthone}{\treewidth}%
\sbox{\treeboxone}{\usebox{\treebox}}\poptree%
\else\ifnum\value{branchcount}=3\sbox{\parentbox}{\ontop{#2}}%
\setlength{\branchwidthone}{\treewidthtwo}%
\addtolength{\branchwidthone}{\treeoffsetone}%
\addtolength{\branchwidthone}{-\treeshiftone}%
\addtolength{\branchwidthone}{-\treeoffsettwo}%
\setlength{\branchwidthtwo}{\treewidththree}%
\addtolength{\branchwidthtwo}{\treeoffsettwo}%
\addtolength{\branchwidthtwo}{-\treeshifttwo}%
\addtolength{\branchwidthtwo}{-\treeoffsetthree}%
\setlength{\branchwidth}{\branchwidthone}%
\ifdim\branchwidthtwo>\branchwidth%
\setlength{\branchwidth}{\branchwidthtwo}\fi%
\setlength{\daughteroffsetone}{\branchwidth}%
\addtolength{\daughteroffsetone}{-\branchwidthone}%
\addtolength{\daughteroffsetone}{-\treeshiftone}%
\setlength{\daughteroffsettwo}{\branchwidth}%
\addtolength{\daughteroffsettwo}{-\branchwidthtwo}%
\addtolength{\daughteroffsettwo}{-\treeshifttwo}%
\setlength{\parentoffset}{-0.5\wd\parentbox}%
\addtolength{\parentoffset}{\treeoffsetthree}%
\addtolength{\parentoffset}{\branchwidth}%
\setlength{\daughteroffset}{0in}%
\ifdim\parentoffset<0in%
\setlength{\daughteroffset}{-\parentoffset}%
\setlength{\parentoffset}{0in}\fi%
\setlength{\parentwidth}{\parentoffset}%
\addtolength{\parentwidth}{\wd\parentbox}%
\setlength{\treeoffset}{\daughteroffset}%
\addtolength{\treeoffset}{\treeoffsetthree}%
\setlength{\treewidth}{\wd\treeboxone}%
\addtolength{\treewidth}{\daughteroffsetone}%
\addtolength{\treewidth}{\treewidthtwo}%
\addtolength{\treewidth}{\daughteroffsettwo}%
\addtolength{\treewidth}{\treewidththree}%
\addtolength{\treewidth}{\daughteroffset}%
\ifdim\treewidth<\parentwidth\setlength{\treewidth}{\parentwidth}\fi%
\sbox{\treebox}{\begin{minipage}{\treewidth}%
\begin{flushleft}%
\hspace*{\parentoffset}\usebox{\parentbox}\\
{\setlength{\unitlength}{0.5\branchwidth}%
\hspace*{\treeoffset}\begin{picture}(4,1)%
\put(0,0){\line(2,1){2}}%
\put(2,0){\line(0,1){1}}%
\put(4,0){\line(-2,1){2}}%
\end{picture}}\\
\vspace{-\baselineskip}
\hspace*{\daughteroffset}%
\makebox[\treewidththree][l]%
{\raisebox{-\ht\treeboxthree}{\usebox{\treeboxthree}}}%
\hspace*{\daughteroffsettwo}%
\makebox[\treewidthtwo][l]%
{\raisebox{-\ht\treeboxtwo}{\usebox{\treeboxtwo}}}%
\hspace*{\daughteroffsetone}%
\raisebox{-\ht\treeboxone}{\usebox{\treeboxone}}%
\end{flushleft}%
\end{minipage}}%
\setlength{\treeoffsetone}{\parentoffset}%
\addtolength{\treeoffsetone}{0.5\wd\parentbox}%
\setlength{\treeshiftone}{0pt}%
\setlength{\treewidthone}{\treewidth}%
\sbox{\treeboxone}{\usebox{\treebox}}\poptree\poptree%
\else\ifnum\value{branchcount}=4\sbox{\parentbox}{\ontop{#2}}%
\setlength{\branchwidthone}{\treewidthtwo}%
\addtolength{\branchwidthone}{\treeoffsetone}%
\addtolength{\branchwidthone}{-\treeshiftone}%
\addtolength{\branchwidthone}{-\treeoffsettwo}%
\setlength{\branchwidthtwo}{\treewidththree}%
\addtolength{\branchwidthtwo}{\treeoffsettwo}%
\addtolength{\branchwidthtwo}{-\treeshifttwo}%
\addtolength{\branchwidthtwo}{-\treeoffsetthree}%
\setlength{\branchwidththree}{\treewidthfour}%
\addtolength{\branchwidththree}{\treeoffsetthree}%
\addtolength{\branchwidththree}{-\treeshiftthree}%
\addtolength{\branchwidththree}{-\treeoffsetfour}%
\setlength{\branchwidth}{\branchwidthone}%
\ifdim\branchwidthtwo>\branchwidth%
\setlength{\branchwidth}{\branchwidthtwo}\fi%
\ifdim\branchwidththree>\branchwidth%
\setlength{\branchwidth}{\branchwidththree}\fi%
\setlength{\daughteroffsetone}{\branchwidth}%
\addtolength{\daughteroffsetone}{-\branchwidthone}%
\addtolength{\daughteroffsetone}{-\treeshiftone}%
\setlength{\daughteroffsettwo}{\branchwidth}%
\addtolength{\daughteroffsettwo}{-\branchwidthtwo}%
\addtolength{\daughteroffsettwo}{-\treeshifttwo}%
\setlength{\daughteroffsetthree}{\branchwidth}%
\addtolength{\daughteroffsetthree}{-\branchwidththree}%
\addtolength{\daughteroffsetthree}{-\treeshiftthree}%
\setlength{\parentoffset}{-0.5\wd\parentbox}%
\addtolength{\parentoffset}{\treeoffsetfour}%
\addtolength{\parentoffset}{1.5\branchwidth}%
\setlength{\daughteroffset}{0in}%
\ifdim\parentoffset<0in%
\setlength{\daughteroffset}{-\parentoffset}%
\setlength{\parentoffset}{0in}\fi%
\setlength{\parentwidth}{\parentoffset}%
\addtolength{\parentwidth}{\wd\parentbox}%
\setlength{\treeoffset}{\daughteroffset}%
\addtolength{\treeoffset}{\treeoffsetfour}%
\setlength{\treewidth}{\wd\treeboxone}%
\addtolength{\treewidth}{\daughteroffsetone}%
\addtolength{\treewidth}{\treewidthtwo}%
\addtolength{\treewidth}{\daughteroffsettwo}%
\addtolength{\treewidth}{\treewidththree}%
\addtolength{\treewidth}{\daughteroffsetthree}%
\addtolength{\treewidth}{\treewidthfour}%
\addtolength{\treewidth}{\daughteroffset}%
\ifdim\treewidth<\parentwidth\setlength{\treewidth}{\parentwidth}\fi%
\sbox{\treebox}{\begin{minipage}{\treewidth}%
\begin{flushleft}%
\hspace*{\parentoffset}\usebox{\parentbox}\\
{\setlength{\unitlength}{0.5\branchwidth}%
\hspace*{\treeoffset}\begin{picture}(6,1)%
\put(0,0){\line(3,1){3}}%
\put(2,0){\line(1,1){1}}%
\put(4,0){\line(-1,1){1}}%
\put(6,0){\line(-3,1){3}}%
\end{picture}}\\
\vspace{-\baselineskip}
\hspace*{\daughteroffset}%
\makebox[\treewidthfour][l]%
{\raisebox{-\ht\treeboxfour}{\usebox{\treeboxfour}}}%
\hspace*{\daughteroffsetthree}%
\makebox[\treewidththree][l]%
{\raisebox{-\ht\treeboxthree}{\usebox{\treeboxthree}}}%
\hspace*{\daughteroffsettwo}%
\makebox[\treewidthtwo][l]%
{\raisebox{-\ht\treeboxtwo}{\usebox{\treeboxtwo}}}%
\hspace*{\daughteroffsetone}%
\raisebox{-\ht\treeboxone}{\usebox{\treeboxone}}%
\end{flushleft}%
\end{minipage}}%
\setlength{\treeoffsetone}{\parentoffset}%
\addtolength{\treeoffsetone}{0.5\wd\parentbox}%
\setlength{\treeshiftone}{0pt}%
\setlength{\treewidthone}{\treewidth}%
\sbox{\treeboxone}{\usebox{\treebox}}\poptree\poptree\poptree%
\else\ifnum\value{branchcount}=5\sbox{\parentbox}{\ontop{#2}}%
\setlength{\branchwidthone}{\treewidthtwo}%
\addtolength{\branchwidthone}{\treeoffsetone}%
\addtolength{\branchwidthone}{-\treeshiftone}%
\addtolength{\branchwidthone}{-\treeoffsettwo}%
\setlength{\branchwidthtwo}{\treewidththree}%
\addtolength{\branchwidthtwo}{\treeoffsettwo}%
\addtolength{\branchwidthtwo}{-\treeshifttwo}%
\addtolength{\branchwidthtwo}{-\treeoffsetthree}%
\setlength{\branchwidththree}{\treewidthfour}%
\addtolength{\branchwidththree}{\treeoffsetthree}%
\addtolength{\branchwidththree}{-\treeshiftthree}%
\addtolength{\branchwidththree}{-\treeoffsetfour}%
\setlength{\branchwidthfour}{\treewidthfive}%
\addtolength{\branchwidthfour}{\treeoffsetfour}%
\addtolength{\branchwidthfour}{-\treeshiftfour}%
\addtolength{\branchwidthfour}{-\treeoffsetfive}%
\setlength{\branchwidth}{\branchwidthone}%
\ifdim\branchwidthtwo>\branchwidth%
\setlength{\branchwidth}{\branchwidthtwo}\fi%
\ifdim\branchwidththree>\branchwidth%
\setlength{\branchwidth}{\branchwidththree}\fi%
\ifdim\branchwidthfour>\branchwidth%
\setlength{\branchwidth}{\branchwidthfour}\fi%
\setlength{\daughteroffsetone}{\branchwidth}%
\addtolength{\daughteroffsetone}{-\branchwidthone}%
\addtolength{\daughteroffsetone}{-\treeshiftone}%
\setlength{\daughteroffsettwo}{\branchwidth}%
\addtolength{\daughteroffsettwo}{-\branchwidthtwo}%
\addtolength{\daughteroffsettwo}{-\treeshifttwo}%
\setlength{\daughteroffsetthree}{\branchwidth}%
\addtolength{\daughteroffsetthree}{-\branchwidththree}%
\addtolength{\daughteroffsetthree}{-\treeshiftthree}%
\setlength{\daughteroffsetfour}{\branchwidth}%
\addtolength{\daughteroffsetfour}{-\branchwidthfour}%
\addtolength{\daughteroffsetfour}{-\treeshiftfour}%
\setlength{\parentoffset}{-0.5\wd\parentbox}%
\addtolength{\parentoffset}{\treeoffsetfive}%
\addtolength{\parentoffset}{2\branchwidth}%
\setlength{\daughteroffset}{0in}%
\ifdim\parentoffset<0in%
\setlength{\daughteroffset}{-\parentoffset}%
\setlength{\parentoffset}{0in}\fi%
\setlength{\parentwidth}{\parentoffset}%
\addtolength{\parentwidth}{\wd\parentbox}%
\setlength{\treeoffset}{\daughteroffset}%
\addtolength{\treeoffset}{\treeoffsetfive}%
\setlength{\treewidth}{\wd\treeboxone}%
\addtolength{\treewidth}{\daughteroffsetone}%
\addtolength{\treewidth}{\treewidthtwo}%
\addtolength{\treewidth}{\daughteroffsettwo}%
\addtolength{\treewidth}{\treewidththree}%
\addtolength{\treewidth}{\daughteroffsetthree}%
\addtolength{\treewidth}{\treewidthfour}%
\addtolength{\treewidth}{\daughteroffsetfour}%
\addtolength{\treewidth}{\treewidthfive}%
\addtolength{\treewidth}{\daughteroffset}%
\ifdim\treewidth<\parentwidth\setlength{\treewidth}{\parentwidth}\fi%
\sbox{\treebox}{\begin{minipage}{\treewidth}%
\begin{flushleft}%
\hspace*{\parentoffset}\usebox{\parentbox}\\
{\setlength{\unitlength}{0.5\branchwidth}%
\hspace*{\treeoffset}\begin{picture}(8,1)%
\put(0,0){\line(4,1){4}}%
\put(2,0){\line(2,1){2}}%
\put(4,0){\line(0,1){1}}%
\put(6,0){\line(-2,1){2}}%
\put(8,0){\line(-4,1){4}}%
\end{picture}}\\
\vspace{-\baselineskip}
\hspace*{\daughteroffset}%
\makebox[\treewidthfive][l]%
{\raisebox{-\ht\treeboxfour}{\usebox{\treeboxfive}}}%
\hspace*{\daughteroffsetfour}%
\makebox[\treewidthfour][l]%
{\raisebox{-\ht\treeboxfour}{\usebox{\treeboxfour}}}%
\hspace*{\daughteroffsetthree}%
\makebox[\treewidththree][l]%
{\raisebox{-\ht\treeboxthree}{\usebox{\treeboxthree}}}%
\hspace*{\daughteroffsettwo}%
\makebox[\treewidthtwo][l]%
{\raisebox{-\ht\treeboxtwo}{\usebox{\treeboxtwo}}}%
\hspace*{\daughteroffsetone}%
\raisebox{-\ht\treeboxone}{\usebox{\treeboxone}}%
\end{flushleft}%
\end{minipage}}%
\setlength{\treeoffsetone}{\parentoffset}%
\addtolength{\treeoffsetone}{0.5\wd\parentbox}%
\setlength{\treeshiftone}{0pt}%
\setlength{\treewidthone}{\treewidth}%
\sbox{\treeboxone}{\usebox{\treebox}}\poptree\poptree\poptree\poptree%
\else\typeout{QobiTeX warning--- Can't handle #1 branching}\fi\fi\fi\fi\fi}
\newcommand{\faketreewidth}[1]{%
\sbox{\parentbox}{\ontop{#1}}%
\setlength{\treewidthone}{0.5\wd\parentbox}%
\addtolength{\treewidthone}{\treeoffsetone}%
\setlength{\treeshiftone}{\treeoffsetone}%
\addtolength{\treeshiftone}{-0.5\wd\parentbox}}
\newcommand{\tree}{%
\usebox{\treeboxone}
\setlength{\treeoffsetone}{\treeoffsettwo}%
\sbox{\treeboxone}{\usebox{\treeboxtwo}}%
\poptree}

\usepackage{epsf}

\title{Modelling Users, Intentions, and Structure in Spoken Dialog}
\author{Bernd Ludwig, G\"unther G\"orz, and Heinrich Niemann}
\begin{document}
\maketitle
\begin{abstract}
We outline how utterances in dialogs can be
interpreted using a partial first order logic. We exploit the
capability of this logic to talk about the truth status of formulae to
define a notion of coherence between utterances and explain how this
coherence relation can serve for the construction of AND/OR trees that
represent the segmentation of the dialog. In a BDI model we
formalize basic assumptions about dialog and cooperative behaviour of
participants. These assumptions provide a basis for inferring
speech acts from coherence relations between utterances and attitudes
of dialog participants. Speech acts prove to be useful for
determining dialog segments defined on the notion of completing
expectations of dialog participants. Finally, we sketch how
explicit segmentation signalled by cue phrases and performatives is
covered by our dialog model.
\end{abstract}
\section{Introduction}
During the last years, a large number of spoken language dialog
systems have been developed whose functionality normally is restricted
to a certain application domain. \cite{SadekMori:97} give an quite
extensive
overview of existing implementations.

Only few systems for generating dialog managers exist or are under
development currently. These tools identify task and discourse
structure and describe it by means of finite state automata. Using
these tools one can easily and quickly implement spoken language
human-machine communication for simple tasks. Nevertheless, this
approach lacks theoretical sufficiency for a large number of phenomena
occurring frequently in natural language dialogs. \cite{SadekMori:97}
state that ``these limitations rule out these
approaches as a basis for computational models of intelligent
interaction''.

Recently, there has been some research on extracting dialog structure
out of annotated corpora (\cite{Moeller:97}); algorithms for learning
probability distributions of speech acts are used in this case. The
estimated distributions serve as a basis for generating stochastic
models for sequences of speech acts. But in this case, exploring
common elements of dialogs in different domains is substituted by an
abstract
optimization process, although knowledge of these elements could be
useful for improving parameter estimation.

On the other hand, many approaches to dialog processing consider
different structural elements important to gain a deeper understanding
of the effects that utterances have on the dialog itself and its
participants. But these approaches do not take spoken language and the
problems related to speech recognition into account.

Following the opinon of \cite{Poesio:94} we consider the
separation of {\it describing} and {\it described situation} to be a
crucial
point for dialog processing. This reflection is backed up by
philological and
linguistic research on discourse (\cite{Diewald:87}, \cite{Martin:92},
\cite{BrietzmannGoerz:82}, \cite{Bunt:97}).

On this basis we present fundamental elements of dialog structure to
handle even spoken language. The main aim of this paper is to
build a bridge between research on spoken language processing and
study of discourse structure and cognitive approaches to communication
between individuals in order to conceptualize dialog systems that
integrate
experience from all areas of research just mentioned.

\section{Different Previous Approaches}
As there exists a vast amount of literature on discourse and user
models, we first give an account of some of the important directions
of research performed up to now. The main interest of all this work is
to make precise the notion of ``context'' which is said to be of great
importance for natural language understanding. Inspired by the work of
Grosz and Sidner (\cite{GroszSidner:86}) on the interrelations of task
and discourse structure, a diversification to structural, semantical and

plan-oriented studies has taken place.

\subsection{Discourse Structure}
The fundamental consideration for work on the structure of discourse
is that there is a correspondence between the ordering of utterances
in a discourse and how they are related among each other on the
semantic level (\cite{Gardent:94}, \cite{Gardent:97}). Tree structures
are used to describe the semantic coherence of discourse. By using
these structures, one can define constraints on possible places for
attaching a new utterance to a existing discourse (\cite{Webber:91})
and on accessibility relations for potential referents for deictic
expressions (\cite{Polanyi:95}). In approaches based on Discourse
Representation Theory (\cite{KampReyle:93}) this
correspondence is captured by construction rules (which can be
defined in terms of an extended $\lambda$-calculus---see
\cite{Kuschert:96}) building up Discourse Representation Structures
that describe the coherence relation of all the included
contributions\footnote{DRT has been extended by many researchers
in different ways---e.g.\ by Asher for capturing discourse segments
(\cite{Asher:93}) or in the VERBMOBIL project for handling complex
phenomena of spontaneous speech for machine translation
(\cite{vm135:96}, \cite{vm83:95}). But all these theories initially
describe monologs and therefore do not consider multi-party
communication which is characteristic for dialogs.}.

\subsection{Speech Act based Theories}
Many semantic theories work only locally, i.e.\ they describe the
meaning of one single utterance, ignoring its context and the
discourse situation in general. Insofar, they are unable to account
for the functionality (i.e.\ intention) of a perceived utterance.

Basing on earlier work by Austin,
\cite{Searle:69} proposed a theory of speech acts that has
been fundamental for research on this area. Speech acts are
implemented in dialog managers to derive hypotheses of how the current
utterance contributes to the dialog so far. But considering the last
utterance only is insufficient for a cooperative dialog participant as
this local view does not pay attention to expectations of other
parties involved in the dialog (e.g.\ when somebody asks a question,
she expects the following utterance to be an answer to
it). Consequently, to describe coherence in dialog steps, the effects
of previous utterances have to be recorded somehow.
So, the structural approach of conversational games
 mentioned briefly above provides an
explanation for the speaker
uttering something in the course of dialog. Another point of view to
the coherence problem is taken by \cite{TraumAllen:94}: they propose
that the speech acts associated to each utterance impose social and
conventional obligations on the hearer and therefore constrain the set
of possible legal responses. Equivalently, one can state that after
uttering something the speaker has certain expectations that she wants
to be fulfilled by any response that will be given in the next dialog
step.

\subsection{Intentions, Plans, and Coherence of Utterances}
By now, we are able to describe how linearly (by time of being
uttered) ordered contributions to a dialog can be integrated into a
(partially ordered) discourse structure, but it is still impossible to
explain the motivations of the speaker to use a certain speech act,
especially in the case when the expectations introduced by the
previous are violated. To answer this question, one has to study the
mental attitudes of dialog participants.

Motivations for engaging in a dialog can be taken into account by
studying planning of utterances. Discourse
planning is discussed extensively e.g.\ by Lambert and Lochbaum
(\cite{Lambert:93} and \cite{Lochbaum:94}). Both authors concentrate on
the integration of domain dependent planning steps into the
interpretation of utterances. Approaches such as those by
\cite{MooreYoung:94} or \cite{CarrollCarberry:94} devise a model for
collaborative plans for response generation.

Another line of research focusses on the BDI model which is a more
domain independent approach. This accentuates an agent-based view of
dialog as the participants in dialog and their personal attitudes are
considered to be of main interest. Beliefs, desires, and intentions
are assumed to drive utterances and speech acts
(\cite{AsherLascarides:97}, \cite{AsherSingh:93}). So the key problem
for the interpretation by the dialog manager is the reconstruction of
the speaker's attitudes from what she uttered. Coherence of utterances
is obtained by fundamental assumptions on cooperative behaviour
(\cite{Grice}) and by analyzing how the content of utterances coheres
on the basis of the (mutually believed) domain knowledge
(\cite{AsherLascarides:91}, \cite{AsherLascarides:94},
\cite{AsherLascaridesOberlander:92}). Along this direction of research
there is also some work on coherence relations between utterances
(\cite{Knott:96}, for an overview see
\cite{BatemanRondhuis:94}). These relations characterize the logical
connection between utterances and thereby serve as a basic instrument
for an analysis of the argumentative structure described by a given
dialog.

\subsection{Spoken Language Phenomena}
In spoken language hesitations, repairs, etc.\ are
very common, because in oral communication concentration on the topic
of the dialog limits mental resources available for speech
production. These phenomena cannot be captured by semantic
formalisms as sketched above. To overcome this problem, multi-level
processing of spoken language utterances has been proposed in the
literature (e.g.\ see \cite{TraumHinkelman:91}).

\section{Interpretation of Utterances}
\label{fil}
This section explains our approach how utterances can be
interpreted using First
Order Partial Information Ionic Logic (FIL, \cite{Abdallah:95}) as
a language for describing the semantics of utterances.

A central issue of dialog management is that dialogs are motivated by
the speaker's desire to add information to the knowledge of the dialog
participants. On the other hand, it occurs frequently that
the {\it shared knowledge} of the participants does not contain enough
information to meet the expectations that are ``pending'' between
speaker and hearer. For that reason, our semantic language must be
able to handle situations of partial knowledge. FIL provides formulae
(so called {\it ionic formulae}) like
$$
\ast (\{\phi_1,...,\phi_k\},\xi)
$$ meaning intuitively that $\xi$ is true when it is plausible that
$\Phi=\{\phi_1,...,\phi_k\}$ (called {\it justification set} or {\it
justification context}) is true, too (see \cite{Abdallah:95}, Sect.\
5). $\Phi$ is the set of missing information to infer $\xi$. FIL can
be used to compute such {\it justification contexts}.

We incorporate FIL for the description of conditions in a DRT-based
framework to represent dialog structure. An example of such a
discourse representation structure (DRS) would be:

\vskip.33\baselineskip
\centerline{{\it Does a plane depart from Athens to Rome?}}
\vskip.33\baselineskip
\noindent has the semantic representation
$$
\left[
\begin{array}{ll}
{\rm t}\,{\rm Rome}\,\mbox{Athens}\\\hline
\mbox{\sc Plane}({\rm t})\\
\mbox{\sc Depart}({\rm t})\\
\mbox{\sc Airport}({\rm Athens})\\
\mbox{\sc Airport}({\rm Rome})\\
\mbox{\sc From}({\rm t},{\rm Athens})\\
\mbox{\sc To}({\rm t},{\rm Rome})
\end{array}
\right]
$$

In DRT, there exists a number of construction rules for incremental
composition of several individual utterances. One can even infer
whether DRS $K_N$ is a consequence of the DRS $K_1$, ...,
$K_{N-1}$. But whilst in standard DRT conditions are described by
classical first order formulae, in our approach FIL is used for that
purpose. As FIL is a partial logic, we can compute whether $K_N$ is
undefined given $K_1$, ..., $K_{N-1}$. This is true, as FIL allows to
talk about the truth value of a formula:
$$
\mbox{undefined}(\phi) \Longleftrightarrow \sim \neg \phi \wedge \sim
\phi
$$
So, we are able to assign one of the following three consequence
states to $K_N$:
\begin{itemize}
\setlength{\parsep}{0pt}
\setlength{\itemsep}{0pt}
\item $\models K_N$. $K_N$ follows from the discourse so far.
\item $\false K_N \Longleftrightarrow \models \neg K_N$: $\neg K_N$
follows from previous utterances.
\item $\models \sim \neg K_N \wedge \sim K_N$: $K_N$ is still undefined.

\end{itemize}
Using the deduction theorem ($\Delta \cup \{\phi_1,...,\phi_K\}\models
\psi
\Longleftrightarrow \Delta \models \phi_1\wedge ...\wedge \phi_K
\rightarrow \psi$) we have
established a coherence relation between utterances via implication in
FIL. When $\phi_1\wedge ...\wedge \phi_K \rightarrow \psi$ is true,
$\ast(\{\phi_1, ..., \phi_K\},\psi)$ is true, too.

Interpreting an utterance requires a knowledge base $\Delta$ for the
representation of the domain relevant knowledge. As described in
\cite{dl98:98}, we use description logics to define the notions to be
understood by the dialog manager for a given application. More
precisely, description logics serve for constructing a terminology of
the domain, thereby representing domain dependent, but situation
independent knowledge. In order to interpret a specific utterance in a
given situation, the given terminology is instantiated by concrete
facts that are entailed by the semantic representation of the current
utterance. To give a simple example of this idea, we could state in
the knowledge base of a flight information system that a flight from a
departure location to an arrival location is a flight characterized by
the existence of an airport at the departure and the arrival location,
respectively. In description logics, we could say:
$$
\mbox{\sc FlightFromTo}=\exists \mbox{\sc From}.\mbox{\sc airport}\cap
\exists
\mbox{\sc To}.\mbox{\sc airport}\cap \mbox{\sc Flight}
$$

So, this definition characterizes knowledge that holds in every
situation in the given application domain. On the other hand, the DRS
above describes a concrete situation.

We conclude that the consequence states mentioned above have to be
understood as consequence on the basis of a situation independent
knowledge base that characterizes the application domain. To interpret
utterances in a given situation, the dialog manager tries to infer the
consequence state of the current utterance relying on his domain
knowledge. In general, this state depends on a certain {\it
justification context} as outlined above. As will be shown below, we
can verify the truth value of all elements in a {\it justification
context} $\Phi$ if we interpret each $\phi_i$ as a question to the
hearer and view the subsequent response as an answer to this
question. This means that the dialog manager's planning steps are
strongly affected by the results of inference in its knowledge
base. From that view and the semantics of $\ast(\{\phi_1, ...,
\phi_K\},\psi)$ we can derive an $n$-ary AND/OR tree that reflects the
discourse structure of the discussed dialog segment.

In the tradition of \cite{GroenendijkStokhof:84} and
\cite{GabbayReyle:92} we consider (free) discourse referents of
interrogative
pronouns as $\lambda$-bound variables. If the problem solver finds a
solution for the posed query, then it binds these variables to
discourse referents that have been introduced earlier (or during the
process of problem solving). Of course, there can be more than one
possible substitution of the $\lambda$-variables. For given variables
$x_1$, ..., $x_M$, we denote a substitution of all variables by
discourse referents $t_1$, ..., $t_M$ as $\Sigma=\{[x_1/t_1], ...,
[x_M/t_M]\}$. $\{\Sigma_1, ..., \Sigma_K\}$ is a set of $K$ pairwise
distinct substitutions.

In the general case where the result of the inference process consists
of a set $\{\Sigma_1, ..., \Sigma_K\}$ of {\it answer substitutions}
and a set $\{\phi_1, ..., \phi_{N-1}\}$ of {\it justification
substitutions}, each $\Sigma_i$ induces
an edge in a OR-subtree of the overall discourse structure.

This tree structure is part of the {\it describing situation} for the
current dialog. Below we will introduce operations on the {\it dialog
tree} that characterize how the structure is expanded in the course of
dialog. In this sense, such a tree constitutes the ``syntax'' of the
current dialog. But there is a strong connection to what could be
called ``dialog semantics''. It is grounded basically on the meaning
of the edges in the tree: they express the fact that parent and child
nodes are coherent in the sense of FIL consequence explained
above. Furthermore, by the notion of satisfiability of FIL ionic
formulae we exploit the tree structure to reformulate Grosz' and
Sidner's relations {\it dominance} and {\it satisfaction
precedence}\footnote{$\alpha$ dominates $\beta$ ($\alpha \uparrow
\beta$) if and only if $\beta$ is part of $\alpha$, while $\alpha$
satisfaction-precedes $\beta$ ($\alpha \prec \beta$) if and only if
$\alpha$ is neccessary for $\beta$}: because {\it justification
context} $\Phi=\{\phi_1,...,\phi_k\}$, when $\ast(\Phi,\psi)$ is
given, is true if and only if all $\phi_i$ are true on the justification

level\footnote{i.e.\ (stated in model-theoretic semantics) there
exists an interpretation that expands any interpretation that makes
$\psi$ true in such a way that it assigns true to all $\phi_i$, too.},
we have $\psi \uparrow \phi_i$ for all $1\leq i\leq k$. And as $\false
\phi_i$ for one $i\in\{1,...,k\}$ implies $\false \ast(\Phi,\psi)$, we
obtain $\phi_i \prec \phi_j$ for $1\leq i < j \leq k$.

\section{Basic Elements of Dialogs}
\subsection{Empirical Evidence for the Need of User Models}

Dialog managers for real world applications have to be robust in the
sense that they always terminate an (user-)initiated dialog in a
controlled
way. So, the study of how to build robust generic dialog managers
implies to reason about what structures exist in a dialog and how they
get modified by utterances. On the other hand, it is also important to
understand how dialogs affect the participants and their future
utterances.

A first approach to this problem is to see the function of utterances
as that of updating the {\it shared knowledge}---a data
structure maintained and used by all dialog participants. From this
point of view, each dialog participant infers the same consequences
from every new utterance.

But as pointed out in the AI literature, actually people hold personal
assumptions about the meaning of an utterance. These assumptions can
differ among dialog participants. We illustrate this by an example
taken from the TRAINS corpus (see \cite{TC93}):

\newcount\dialogstep
\def\begindialog{\vskip.5\baselineskip \par \dialogstep=1
\setbox100=\vbox
\bgroup \hsize=.49\hsize}
\def\enddialog{\egroup \bgroup \splittopskip=0pt \vbadness=10000
\dimen1=\pagegoal \advance \dimen1 by -\pagetotal
\ifdim\dimen1<0pt \advance \dimen1 by\vsize\fi
\ifdim.5\ht100>\dimen1 \setbox2=\vsplit100 to\dimen1
\setbox3=\vsplit100 to\dimen1
\dimen1=\ht2
\hbox to\hsize{\vbox to\dimen1{\unvbox2\vss}\hss \vrule \hss \vbox
to\dimen1{\unvbox3 \vss}}%
\setbox2=\vsplit100 to.5\ht100
\dimen1=\ht2
\hbox to\hsize{\vbox to\dimen1{\unvbox2\vss}\hss \vrule \hss
\vbox to\dimen1{\unvbox100\vss}}
\else \global\setbox101=\vsplit 100 to .5\ht100
 \hbox to\hsize{\copy101 \hss \vrule \hss \vbox to
\ht101{\unvbox 100\vss}}\fi \egroup
\vskip\baselineskip}

\begindialog
\def\utt#1{\hangafter=1 \hangindent=.75cm \raggedright #1\par}
\utt{1.1 {\bf M}: okay}
\utt{1.2  : we have to get .. a}
\utt{1.3  : tanker car of orange juice to uh Avon}
\utt{1.4  : and a}
\utt{1.5  : boxcar of bananas to Corning}
\utt{1.6  : and we have to do that by 3 PM today}
\utt{2.1 {\bf S}: okay}
\utt{3.1 {\bf M}: okay}
\utt{3.2  : so let's see umm}
\utt{3.3  : ... we probably have to take the tanker car}
\utt{3.4  : from Corning to Elmira}
\utt{3.5  : to get uh}
\utt{3.6  : orange juice in it}
\utt{3.7  : um}
\utt{  [2sec] }
\utt{3.8  : [click] and uh}
\utt{3.9  : how far is it from Corning to Elmira}
\utt{3.10  : how long would it take}
\utt{4.1 {\bf S}: 2 hours}
\utt{5.1 {\bf M}: m hm}
\utt{5.2  : okay so}
\utt{  [2sec]}
\utt{5.3  : why don't we uh}
\utt{5.4  : let's see}
\utt{5.5  : [sniff]}
\utt{5.6  : okay why don't we}
\utt{5.7  : would w/ }
\utt{5.8  : uh}
\utt{5.9  : why don't we consider sending uh}
\utt{5.10  : engine E2}
\utt{5.11  : to Corning}
\utt{5.12  : to get the tanker car}
\utt{5.13  : and uh}
\utt{5.14  : bring it back to Elmira}
\utt{  [2sec]}
\utt{5.15  : uh}
\utt{5.16  : and uh have them}
\utt{  [2sec]}
\utt{5.17  : have them fill it}
\utt{5.18  : with OJ}
\utt{5.19  : so how long would it take}
\utt{6.1 {\bf S}: well y  / you need to get oranges}
\utt{6.2  : to the OJ factory}
\utt{7.1 {\bf M}:              +oh + okay}
\utt{7.2  : there's no oranges there yet}
\utt{7.3  : okay so}
\enddialog

In (1.1) to (1.6) {\bf M} describes the goal of this dialog and some
constraints. Doing that he makes some of his mental attitudes public,
thereby assuming that {\bf S} will be able to interpret them
appropriately. So (1.1) to (1.6) do not transport content about the
domain, but about {\bf M} exclusively. The impact of the observation
that
utterances can contain domain relevant knowledge as well as knowledge
about other dialog participants is enormous: In (6.1) and (6.2) {\bf S}
tries to explain why {\bf M} will not be able to reach his goals by
explaining why {\bf M}'s information about the domain and the current
domain
scenario is incomplete or false. This is possible only because {\bf S}
can
differentiate between his own domain knowledge and that transported by
the previous utterances. As a consequence, {\bf S}' cooperative
behaviour is
made possible by his ability to reason about the domain and about his
assumptions of {\bf M}'s view of the domain.

This example shows that cooperative dialog managers must maintain some
sort of user model. Our approach will be discussed in the remainder of
this section.

\subsection{Rational Behaviour of Dialog Participants}
\label{usermodel}
Discussing dialog management, one normally assumes that people
initiate communication with others in order to get help for achieving a
certain goal. From these observations, we can
derive that questions are asked to get an answer that completes the
speaker's knowledge in some way.

For an answer to be helpful, it has to meet certain constraints
(expressed by \cite{Grice} in his maxims of cooperation):
first, it has to be coherent with the question so that it can deliver
valuable information. And, of course, it should be true. These
requirements pose constraints on the behaviour of the person that is
giving the reponse, too. This person has to be cooperative, i.e.\ she
should adopt the speaker's goals, as far as she can realize
them. Honesty is another crucial point. For a person not feeling
obliged to telling the truth is not a reliable source of information.

Our dicussion is restricted to dialogs that fulfill the
requirements above. I.e., we assume some amount of rational behaviour
for all dialog participants.

To reason about goals and intentions, one has to study the cognitive
structures that underly rational behaviour.
For dialogs, these structures are described at length in
\cite{TR663:96}.

One major challenge for designing robust dialog systems seems to be
how to reconstruct the contents of the mental states of the user out
of the utterances---the only observable facts. So the study of how
language can transfer attitudes and reflect planning steps of dialog
participants becomes very important. \cite{Engel:88} notes that in
German basically one can communicate three different types of
utterances:
\begin{itemize}
\setlength{\parsep}{0pt}
\setlength{\itemsep}{0pt}
\item constative: state facts. Our requirement of honesty admits the
conclusion that nothing wrong is intentionally stated to be true.
\item interrogative: ask something.
\item imperative: request something.
\end{itemize}

\subsection{Discourse Domain and Application Domain}
\label{structure}
To formalize the intuitions described up to now, we need a framework
that enables us to talk about utterances or the corresponding DRS,
respectively.

We can achieve this by
introducing a {\it discourse domain}\footnote{The discourse domain is
{\it not} the domain of discourse as we try to make clear in the
remainder of this section. We call the domain of discourse the {\it
application domain} to express the fact that discourse structure and
application structure (sometimes called task structure) are different
and not isomorphic.} that is the domain of {\it describing
situations}. Atomic elements are DRS whereas in the application domain
(consequently to be defined as the domain of {\it described
situations}) atomic
elements are objects of the application. This reflects the ability of
natural language to ``climb up'' to a meta-level, e.g.\ simply by
saying ``What you have told me up to now, has been clear to me. But I
can't understand what I should do now.'' If one thinks about a situation

when a teacher instructs a pupil how to achieve some goal, then it
becomes obvious that responses as above refer only to the
instructions, but not to what is instructed currently.
We use this framework to reason about the relation of utterances and
speaker's attitudes. Following the notation in
\cite{AsherLascarides:97}, we express the connection between
formulating attitudes and the attitudes themselves in the following
way\footnote{We denote the speaker $I$ (initiator) and the hearer $R$
(responder). $\cal B$ expresses beliefs, $\cal W$ desires, $\cal I$
intentions, and $\cal K$ knowledge}:
\begin{eqnarray}
\constative{K}&\rightarrow&\des I{\bel R{K}}\\
\interrogative{K}&\rightarrow&\des I{\know I{K}}\\
\imperative{K}&\rightarrow&\des{I}{\action{R}{K}}
\end{eqnarray}
To describe defeasible rules, we exploit FIL ionic formulae: defaults
eventually to be defeated by overriding exceptions are expressed as
ionic formulae. By that the consequence of an implication is assumed
to be valid as long as there is no evidence to the contrary. If we
have $\phi\rightarrow \ast(\psi,\psi)$ and $\phi$, then $\psi$ is
considered true, unless there is evidence for $\psi$ to be false.
So we can formulate
\begin{itemize}
\setlength{\parsep}{0pt}
\setlength{\itemsep}{0pt}
\item a principle of sincerity:
\begin{eqnarray}
\int I{\bel R{K}}&\rightarrow & \ast(\bel I{K},\bel I{K})
\end{eqnarray}
\item principles of cooperation:
\begin{eqnarray}
\des I{K}\wedge \bel I{\neg K} &\rightarrow & \ast(\des R{K} \wedge
\bel R{\neg K},\des R{K} \wedge \bel R{\neg K})\\
\des I{\action I{K}} &\rightarrow & \ast(\des R{\know I{K}},\des R{\know
I{K}})
\end{eqnarray}
\end{itemize}

After having defined how linguistically motivated types of utterances
and mental states are interrelated among each other and how
fundamental principles of collaboration in dialogs are expressed
formally, we show (by means of two examples) how mental states and
speech acts
are connected via defeasible rules in FIL:
\begin{eqnarray}
\des I{\know I{K}} & \rightarrow & \ast(\query I{K},\query I{K})\\
\nonumber \bel R{\neg(\sim \neg (\xi\rightarrow K)
\wedge \sim (\xi \rightarrow K))}\wedge\\
\label{inform-rule}\des R{\know I{K}}\wedge \des R{\bel I{\xi}}&
\rightarrow & \ast(\inform R{\xi},\inform R{\xi})
\end{eqnarray}
Informally, the first rule states that anything one wants to know is
normally asked for, while the second rule claims that if one wants
another dialog participant to know something and believes it could be
a consequence of something else and wants the other to know that, too,
then
one informs about that new fact.

We can draw the
following conclusion: After the speaker ($I$) has asked $K$ and gets
$\xi$ as response, she can infer defeasibly $\inform R{\xi}$ assuming
that $\xi$ is constative and relying on cooperation if there is no
evidence for $R$ (the hearer of $I$'s utterance, but now speaker of
$\xi$) against the belief $\xi \rightarrow K$.
Furthermore, by sincerity $I$ can also infer that $R$ intends $I$ to
believe $\xi \rightarrow K$.

This example outlines our approach of to how reconstruct speech acts
from
observable utterance types via reasoning about the speaker's
intentions. Recognizing speech acts is important for a cooperative
dialog manager as it helps to explore what state the dialog
currently is in (e.g.\ a question has been asked and an appropriate
answer is now expected). It seems to be just this notion of dialog
state that captures the interactive nature of dialogs in contrast to
monological discourse like a newspaper article. Describing how the
transition of the dialog state looks like in the course of utterances
therefore extends the notion of coherence in discourse sketched
above. But, as can be seen easily from the {\sf inform} rule
(\ref{inform-rule}) above, coherence
is the link betwen mental states and speech acts: no speech act can be
recognized without reasoning about coherence of utterances. In the
rule above, $\inform R{\xi}$ can be inferred only if $\models \xi
\rightarrow K$ or $\false \xi \rightarrow K$. The case when $\xi
\rightarrow K$ is undefined, will be discussed later.

\subsection{Dialog States and Dialog Structure}
First we turn to the discussion of dialog states and their
transition relation: As noted in the literature (see e.g.\
\cite{TraumAllen:94, CristeaWebber:97}), utterances in dialogs induce
expectations of what a plausible response should look like. Stated
differently, by her speech act, the speaker poses some obligation on
the hearer that constrains the preferred responses. Nevertheless, a
model of dialog that does not restrict the range of ``valid''
utterances too much, should give an account for a cooperative reaction
even if
expectations have been violated in a response.

Meeting an expectation and thereby completing the conversational game
opened by the initator, the responder implicitly performs a
segmentation of the discourse. In what follows, we describe the syntax
of discourse segments using a context free grammar whose rules are
attributed by operations on the dialog structure depending on the
coherence between the current utterance and the knowledge shared
between the dialog participants. In that way, we define the semantics
of discourse segmentation in terms of coherence of utterances using
the notion of the AND/OR tree introduced above.

\subsubsection{Simple Segments}
In a simple segment the speaker poses no obligations on the
hearer. When no obligations are pending, an {\sf inform} act would
create a
simple segment according to the following rule:
\begin{eqnarray}
\mbox{SEG}&\rightarrow& \inform I{K}\\
\nonumber && \mbox{SEG}=\newand{\mbox{SEG},K}
\end{eqnarray}
This rule states $K$ can be added as a new fact to the {\it shared
knowledge}, and therefore be inserted as an AND branch into the
currently centered segment. In the absence of pending expectations, the
root segment (i.e.\ the root of the dialog tree) is centered by
default\footnote{Below, after having introduced the operations on the
dialog tree, we will give an example illustrating the application of
the segmentation rules.}.

\subsubsection{Complex Segments}
Dialog segments are called complex when the speaker assigns
expectations to her utterance. For example, when asking a question she
expects an answer to be given by the hearer that helps perform her
intentions. But it is obvious that the hearer could have reason not to
fulfill the expectations (at least temporarily): she could need
additional information first in order to answer the question and
therefore respond with a new one. After this question will have
been answered the hearer will respond according to the still pending
expectation. We capture this situation with the following rules:
\begin{eqnarray}
\mbox{SEG} &\rightarrow & \query I{K}\, \mbox{QUERYEXP}\\
\nonumber &&\mbox{SEG}_1 =
\newand{\mbox{SEG}_1,\newsubtree{K,\mbox{QUERYEXP}}}\\
\mbox{SEG} &\rightarrow & \mbox{SEG}\,\mbox{SEG}\\
\nonumber &&\mbox{SEG}_1=\newand{\mbox{SEG}_2,\mbox{SEG}_3}
\end{eqnarray}
Here, the first rule says that a segment can be initiated by a query
imposing an obligation to answer it. On the other hand, as it is
expressed in the second rule, a segment can consist of two subsegments
forming an AND subtree. QUERYEXP follows the rule:
\begin{eqnarray}
\mbox{QUERYEXP}& \rightarrow & \mbox{ANSWER}\\
\nonumber && \mbox{QUERYEXP}=\mbox{ANSWER}\\
\mbox{QUERYEXP} & \rightarrow & \mbox{ANSWER}\, \mbox{QUERYEXP}\\
\nonumber &&\mbox{QUERYEXP}_1=\newor{\mbox{ANSWER}_2,\mbox{QUERYEXP}_3}
\end{eqnarray}
ANSWER is defined as follows:
\begin{eqnarray}
\mbox{ANSWER} & \rightarrow & \inform I{K}\\
\nonumber &&\mbox{ANSWER}=\newnode{K}\\
\mbox{ANSWER} & \rightarrow & \mbox{SEG}\,\inform I{K}\\
\nonumber &&\mbox{ANSWER}=\changeroot{\mbox{SEG},K}
\end{eqnarray}

\subsection{Example}
\label{example}
In this section, we give an example of how our dialog model behaves at
work. For the purpose of illustration we use the following short flight
information dialog:
\vskip\baselineskip
\user{$\alpha$}{I want a flight from Athens.}
\system{$\beta_1$}{Where do you want to go to?}
\user{$\gamma_1$}{To Rome.}
\system{$\beta_2$}{On which day?}
\user{$\gamma_2$}{On Monday.}
\system{$\beta_3$}{When do you want to depart?}
\user{$\gamma_3$}{At 12 o'clock.}
\system{$\rho$}{There are several connections:
AZ717 at 06:55,
OA233 at 09:05,
AZ725 at 09:20,
OA235 at 12:55,
AZ721 at 16:00,
OA239 at 17:10,
AZ723 at 20:20.}
\vskip\baselineskip
As will be discussed in Sect.\ \ref{performatives}, {\it I want}
expresses a desire explicitly and is therefore represented as $\des I
\alpha$. From that we conclude $\ast(\query I \alpha,\query I
\alpha)$. Besides of that, we have the implications $\des R{\know I
\alpha}$ by cooperation and $\neg \inform I\alpha$. On the other hand,
grammatically $\alpha$ is constative. Therefore
$\mbox{constative}(\alpha)\rightarrow \des I{\know R \alpha}$ and,
consequently, $\ast(\inform I\alpha,\inform I \alpha)$. But
justification
context $\inform I \alpha$ is in conflict with $\neg \inform
I\alpha$. Therefore, the only derivable hypothesis for $\alpha$'s
speech act is $\query I\alpha$. According to the rules for discourse
segmentation above, this induces a QUERYEXP for $R$ (i.e.\ the dialog
manager). This obligation is expressed in the user model by $\des
R{\know I \alpha}$.

In order to fulfill this desire, the dialog
manager tries to evalute the consequence status of $\alpha$ (see
Sect.\ \ref{fil}). This evaluation is performed on the basis of the
 domain model (for a more elaborate example see
\cite{dl98:98}). Its result is that the status of $\alpha$ depends on
the justification context $\{\mbox{\sc At}, \mbox{\sc To}, \mbox{\sc
On}\}$. I.e.\ departure day and time as well as the destination are
still unknown.

So, mental attitudes of the dialog manager change: a
new desire $\des R{\know R{\beta_1}}$ is added that implies
$\ast(\query R{\beta_1},\query R{\beta_1})$. I.e.\ a new discourse
obligation is being introduced leaving the old one pending.

For the user's response $\mbox{constative}(\gamma_1)$ holds, implying
$\des I{\know R {\gamma_1}}$. From $\gamma_1$ the dialog manager infers
$\gamma_1 \rightarrow \beta_1$. Consequently, it assumes $\bel
I{\gamma_1\rightarrow \beta_1}$ and $\ast(\inform I
{\gamma_1},\inform I{\gamma_1})$. As there is no evidence against
$\inform I{\gamma_1}$ and no other hypothesis for a possible speech
act, $\inform I{\gamma_1}$ is assumed. This completes $\query
R{\beta_1}$. With this completion, the dialog manager obtains
information about the destination.

Processing of $\beta_2$, $\gamma_2$ and $\beta_3$, $\gamma_3$ is
performed analogously. After that, the justification context inferred
during the interpretation $\alpha$ has been answered completely. Having
got
all this information the dialog manager can compute a solution for
$\alpha$ that is uttered in $\rho$.

\begin{figure}[ht]
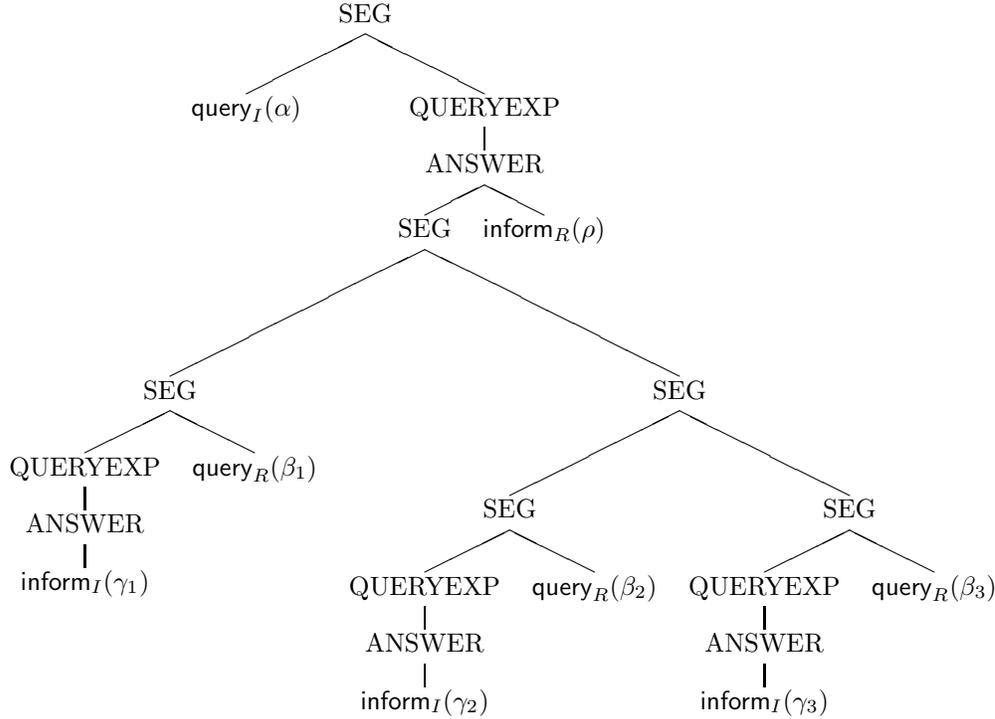

\leaf{$\query I{\alpha}$}
\leaf{$\inform I{\gamma_1}$}
\branch{1}{ANSWER}
\branch{1}{QUERYEXP}
\leaf{$\query R{\beta_1}$}
\branch{2}{SEG}
\leaf{$\inform I{\gamma_2}$}
\branch{1}{ANSWER}
\branch{1}{QUERYEXP}
\leaf{$\query R{\beta_2}$}
\branch{2}{SEG}
\leaf{$\inform I{\gamma_3}$}
\branch{1}{ANSWER}
\branch{1}{QUERYEXP}
\leaf{$\query R{\beta_3}$}
\branch{2}{SEG}
\branch{2}{SEG}
\branch{2}{SEG}
\faketreewidth{SEG}
\leaf{$\inform R{\rho}$}
\branch{2}{ANSWER}
\branch{1}{QUERYEXP}
\faketreewidth{QUERYEXPQUERYEXP}
\branch{2}{SEG}
\setbox1=\hbox{\tree}
\setbox2=\hbox{QQUERYEXP}
\hbox to\textwidth{\hskip\wd2\box1\hss}
\caption{\label{exstructure}Segmentation of the Example Dialog}
\end{figure}

\begin{figure}[ht]
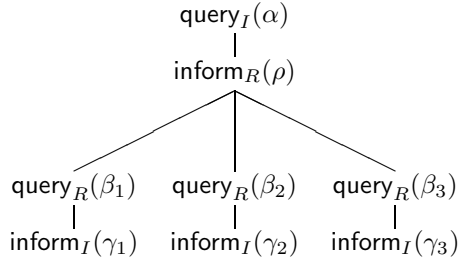

\leaf{$\inform I{\gamma_1}$}
\branch{1}{$\query R{\beta_1}$}
\leaf{$\inform I{\gamma_2}$}
\branch{1}{$\query R{\beta_2}$}
\leaf{$\inform I{\gamma_3}$}
\branch{1}{$\query R{\beta_3}$}
\branch{3}{$\inform R{\rho}$}
\branch{1}{$\query I{\alpha}$}
\centerline{\tree}
\caption{\label{semantics}Coherence structure of the Example Dialog}
\end{figure}

\begin{figure}[ht]
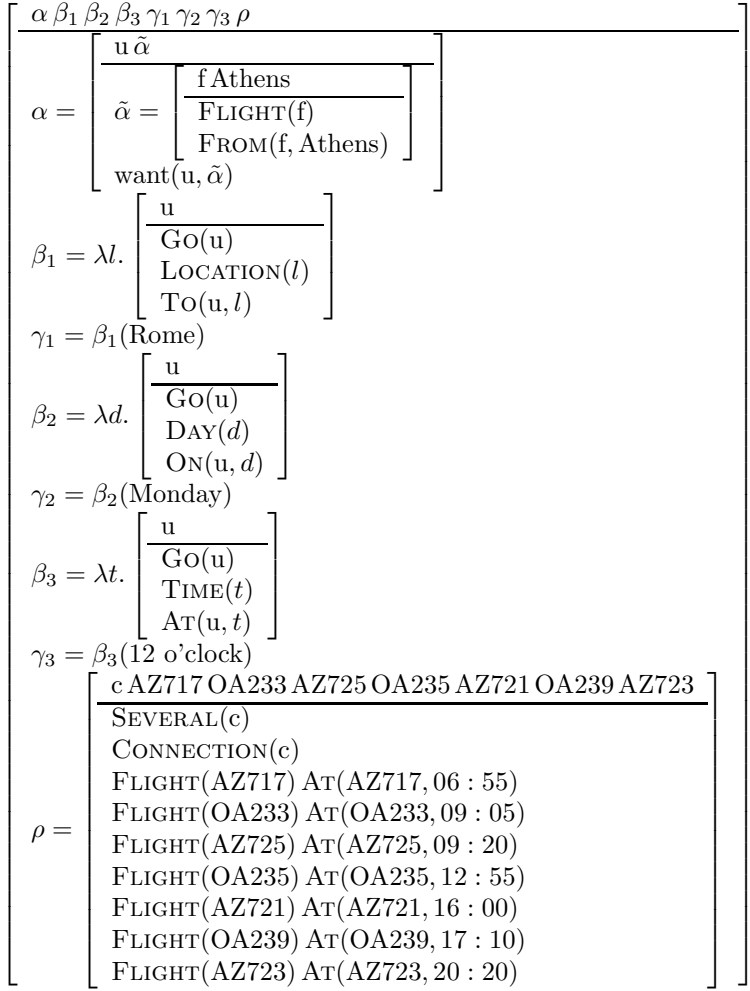

$$
\left[
\begin{array}{l}
\alpha\,\beta_1\,\beta_2\,\beta_3\,\gamma_1\,\gamma_2\,\gamma_3\,\rho\\\hline

\alpha=\left[
\begin{array}{l}
\mbox{u}\,\tilde{\alpha}\\\hline
\tilde{\alpha}=\left[
\begin{array}{l}
\mbox{f}\,\mbox{Athens}\\\hline
\mbox{\sc Flight}(\mbox{f})\\
\mbox{\sc From}(\mbox{f},\mbox{Athens})
\end{array}
\right]\\
\mbox{want}(\mbox{u},\tilde{\alpha})
\end{array}
\right]\\
\beta_1=\lambda l.\left[
\begin{array}{l}
\mbox{u}\\\hline
\mbox{\sc Go}(\mbox{u})\\
\mbox{\sc Location}(l)\\
\mbox{\sc To}(\mbox{u},l)
\end{array}
\right]\\
\gamma_1=\beta_1(\mbox{Rome})\\
\beta_2=\lambda d.\left[
\begin{array}{l}
\mbox{u}\\\hline
\mbox{\sc Go}(\mbox{u})\\
\mbox{\sc Day}(d)\\
\mbox{\sc On}(\mbox{u},d)
\end{array}
\right]\\
\gamma_2=\beta_2(\mbox{Monday})\\
\beta_3=\lambda t.\left[
\begin{array}{l}
\mbox{u}\\\hline
\mbox{\sc Go}(\mbox{u})\\
\mbox{\sc Time}(t)\\
\mbox{\sc At}(\mbox{u},t)
\end{array}
\right]\\
\gamma_3=\beta_3(\mbox{12 o'clock})\\
\rho=\left[
\begin{array}{l}
\mbox{c}\,
\mbox{AZ717}\,
\mbox{OA233}\,
\mbox{AZ725}\,
\mbox{OA235}\,
\mbox{AZ721}\,
\mbox{OA239}\,
\mbox{AZ723}\\\hline
\mbox{\sc Several}(\mbox{c})\\
\mbox{\sc Connection}(\mbox{c})\\
\mbox{\sc Flight}(\mbox{AZ717})\, \mbox{\sc At}(\mbox{AZ717},06:55)\\
\mbox{\sc Flight}(\mbox{OA233})\, \mbox{\sc At}(\mbox{OA233},09:05)\\
\mbox{\sc Flight}(\mbox{AZ725})\, \mbox{\sc At}(\mbox{AZ725},09:20)\\
\mbox{\sc Flight}(\mbox{OA235})\, \mbox{\sc At}(\mbox{OA235},12:55)\\
\mbox{\sc Flight}(\mbox{AZ721})\, \mbox{\sc At}(\mbox{AZ721},16:00)\\
\mbox{\sc Flight}(\mbox{OA239})\, \mbox{\sc At}(\mbox{OA239},17:10)\\
\mbox{\sc Flight}(\mbox{AZ723})\, \mbox{\sc At}(\mbox{AZ723},20:20)
\end{array}
\right]
\end{array}
\right]
$$
\caption{\label{drssem}Discourse Representation Structure for the
Example Dialog}
\end{figure}

Fig.\ \ref{exstructure} shows the dialog segmentation for this example,
while Fig.\ \ref{semantics} sketches the coherence structure of all
utterances, represented in a DRS in Fig.\ \ref{drssem}. A point
worth mentioning is that the structure of the justification context as
an AND tree is embedded in the coherence structure of the dialog.

\section{Incoherence and its Effects on Dialog Structure}
This section explains in more detail how our dialog manager handles
situations when two utterances are incoherent in the sense mentioned
earlier although they otherwise would not violate any
expectations. Consider the dialog is Fig.\ \ref{exampledialog} for a
motivation of the problem:
\begin{figure}[htb]
\begin{tabular}{llp{.75\textwidth}}
?$\alpha$ & {\bf U}: & Is there a flight to Rome on Saturday?\\
.$\beta$ & {\bf S}: & Yes. LH745 at 10:38, AZ304 at 15:03, or 2G261 at
16:25.\\
?$\gamma$ & {\bf S}: & Which airline do you prefer?\\
.$\delta$ & {\bf U}: & The Alitalia flight would be quite convenient.\\
?$\epsilon$ & {\bf U}: & Do they offer business class?\\
.$\zeta$ & {\bf S}: & Yes, they do.\\
?$\eta$ & {\bf S}: & Have you got a MilleMiglia card?\\
.$\theta$ & {\bf U}: & [Hmm, actually] I rather prefer Lufthansa due to
their
superb service.
\end{tabular}
\caption{\label{exampledialog}Example Dialog}
\end{figure}

This dialog is quite regular according to the rules on
dialog state described in the previous section until {\bf U} utters
$\theta$. Shared domain knowledge $\Delta$ and the facts collected
during the
dialog do not allow to conclude that $\theta$ and $\eta$ are
coherent. One can derive that $\Delta \cup \{\theta\} \models
\sim \neg \eta \wedge \sim \eta$. So the question arises which speech
act to assign to $\theta$. To answer that question we have to go back
some steps in the dialog: in occasion of uttering $\gamma$, {\bf S} made
{\bf U}
assume $\des S{\know S{\gamma}}$ implying $\des U{\know
S{\gamma}}$. From this observation and from $\Delta \cup \{\zeta\}
\models \gamma$, {\bf S} can infer that {\bf U} wants to give an answer
to
$\gamma$ ---thereby ignoring {\bf S}'s expectation that $\epsilon$ will
be
answered by {\bf U}.

How can such a situation be described in our model of dialog states?
Following \cite{Traum:94}, we assume that several speech acts can be
assigned to one utterance allowing the speaker to express multiple
intentions at one time. Except of the {\sf inform} act derived be
inferring
coherence between $\theta$ and $\gamma$, we assign a {\sf cancel} act to
{\bf U}'s
last utterance. {\sf cancel} has the meaning that a pending dialog
obligation is violated intentionally as it is the case when {\bf U}
referred to
$\gamma$ when responding to $\eta$.

{\sf cancel} is defined by
\begin{eqnarray}
(\sim \neg K \wedge \sim K)\wedge \inform R{K} & \rightarrow & \cancel
R{K}
\end{eqnarray}
To integrate {\sf cancel} into speech act processing, we add a new rule
for
ANSWER:
\begin{eqnarray}
\mbox{ANSWER} & \rightarrow & \cancel I{K}\\
&& \mbox{ANSWER} = \emptyset
\end{eqnarray}

\section{Configuration of Dialog Managers}
Our approach to dialog understanding as it has been characterized up
to now is dominated by the idea to separate domain independent
algorithms and data structures from domain dependent data for specific
applications and to separate the discourse domain proper from the
application
domain. By the isolation of domain independent dialog elements we try
to explore the minimal amount of dialog structures to be configured
for a specific application. In particular, we have distinguished three
models contributing to the setup of a dialog system for a given
application:

\begin{itemize}
\item Domain model

It defines the notions that exist in the application domain and how
these notions are being interpreted. Additionally, it describes how
the vocabulary of the domain is connected with notions defined in the
domain model (see Sect.\ \ref{fil}).

\item Dialog model

The conversational games valid for the application (see Sect.\
\ref{structure}) and the rules how moves of different games can be
interleaved
among each other are defined in the dialog model. Rules
for the games specify how the dialog structure (represented
by the dialog manager as an AND/OR tree) is affected by a certain
game. In addition, it is possible to restrict dialog participants to
different
sets of conversational games that they are allowed to begin. E.g.\ in
a dialog model without mixed initiative the user would not be
permitted to begin a {\sf query} game, but only be allowed to react
with {\sf inform}. It is an open question whether a restriction of the
kind just described could serve as a sufficient characterization of the
complexity of dialogs.

\item Model of dialog participants (user model)

In order to reason about motivations for conversational games one has
to connect game moves (i.e.\ speech acts) with mental attitudes of
dialog participants. For this purpose, the user model defines
neccessary conditions of mental attitudes for each speech act. On the
other hand, it also contains the general principles of rational
behaviour that hold between the attitudes of dialog participants (see
\ref{usermodel}).
\end{itemize}

During the configuration for a specific application the models
sketched above have to be defined. We argue that at least for the
dialog model and the user model there exists a large application
independent subset of definitions that holds for any application and
has only to be completed for a concrete domain.

In many cases, such a subset would reduce configuration to the
definition of an appropriate domain model. From this point of view, it
would be worth analyzing which classes of dialogs could be covered by
proposals for domain independent sets of speech acts (such as the one
described by the Discourse Resource Initiative---see \cite{DRI}).

\section{Explicit Modification and Segmentation of Dialog Structures}
Any model for describing discourse as the one sketched in this paper
should give an account not only for an implicit construction of dialog
structures, but also for its explicit modification by the dialog
participants (see e.g.\ \cite{Cohen:90}). Such an account would reflect
the capability of dialog to talk about what
\cite{Bunt:97} has called ``dialog control'', or about attitudes and
mental states of dialog participants. Switching to the ``meta-level''
is normally signalled by cue phrases or performatives.

In the remainder of this section we will discuss how these special
types of utterances that have a well-defined meaning only in the {\it
describing situation} of the dialog affect our dialog model.

\subsection{Cue Phrases}
In our opinion, many natural language expressions are like
polymorphous operators in object-oriented programming languages: they
take arguments of different type and have different semantics each
time. This view is shared by other researchers, too---e.g.\
\cite{vanBenthem:88}. E.g.\ in the utterance ``Do you want to
depart from Munich or from Frankfurt?'' {\it or} expresses a choice
between two locations---i.e.\ two objects of the {\it described
situation}. On the other hand, in ``Will you go there by bus or rather
take the car?'' {\it or} again states two possible alternatives, but,
in this case, they are utterances---i.e.\ objects of the {\it
describing situation}.

%\marginpar{\tiny vom Ende von 7.1 hierher}
To incorporate interpretation of cue phrases of into the
dialog model
we rely on Knott's work (\cite{Knott:96}) on coherence
relations. Knott discusses extensively how cue phrases contribute to
the understanding of discourse coherence: He assumes that any cue
phrase has the function of an operator between previous utterances
$\alpha_1$, ..., $\alpha_N$ and an utterance $\beta$ following the cue
phrase. These utterances are connected by a defeasible rule
$P_1\wedge...\wedge P_N\rightarrow C$ which we can
express in FIL as $\ast(P_1\wedge...\wedge P_N\rightarrow
C,P_1\wedge...\wedge P_N\rightarrow C)$. Each cue
phrase has an associated set of features like polarity of the
consequent etc.\ that define its semantics and how the utterances in
the scope of the cue phrase are linked to $P_i$ and $C$. E.g.\ for
{\it but}, we have $P_i:=\neg \alpha_i$ and $C:=\beta$. Consequently
for ``$\alpha$, {\it but} $\beta$'' coherence between $\alpha$ and
$\beta$ is expressed by $\neg \alpha \rightarrow \beta$.

For the general case, coherence between utterances in the scope of the
cue phrase can be established if $\{P_1, ...,P_N\}\models C$ or
$\{P_1, ...,P_N\}\false C$. This result can be exploited to update the
dialog structure appropriately.

For a deeper investigation of this topic let's have a look at the
following dialog:
\vskip.5\baselineskip
\begin{tabular}{llp{.75\textwidth}}
?$\alpha$ & {\bf U}: & Is there a flight to Rome on Saturday?\\
.$\beta$ & {\bf S}: & Yes. AZ631 at 15:03\\
?$\gamma$ & {\bf U}: & How much is it?\\
.$\delta$ & {\bf S}: & DM 528 plus tax.\\
?$\epsilon$ & {\bf U}: & {\it or} on Monday, what about that?\\
.$\zeta$ & {\bf S}: & On Monday you can fly with Debonair.\\
?$\eta$ & {\bf U}: & How much is a ticket?\\
.$\theta$ & {\bf S}: & DM 199 plus tax.\\
\end{tabular}
\vskip.5\baselineskip
How is ``{\it or} on Monday?'' processed and integrated into the
dialog structure? Firstly, it has to be noticed that ellipsis
resolution on the PP ``on Monday'' yields as syntactic referent ``Is
there a flight to Rome?''.

\begin{figure}[htb]
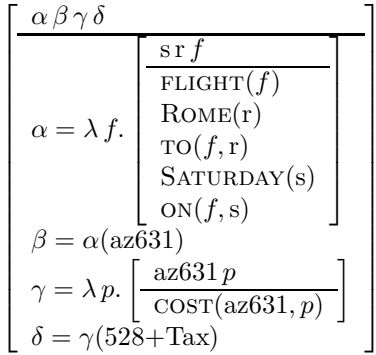

$$
\left[
\begin{array}{l}
\alpha\,\beta\,\gamma\,\delta\\\hline
\alpha=\lambda\,f.
\left[
\begin{array}{l}
\mbox{s}\,\mbox{r}\,f\\\hline
\mbox{\sc flight}(f)\\
\mbox{\sc Rome}(\mbox{r})\\
\mbox{\sc to}(f,\mbox{r})\\
\mbox{\sc Saturday}(\mbox{s})\\
\mbox{\sc on}(f,\mbox{s})
\end{array}
\right]\\
\beta=\alpha(\mbox{az631})\\
\gamma=\lambda\,p.
\left[
\begin{array}{l}
\mbox{az631}\,p\\\hline
\mbox{\sc cost}(\mbox{az631},p)
\end{array}
\right]\\
\delta=\gamma(\mbox{528+Tax})
\end{array}
\right]
$$
\caption{\label{drs1}DRS after $\delta$ has been processed.}
\end{figure}

Secondly, the DRS for the {\it describing situation} after $\delta$ has
been
processed, is as shown in Fig.\ \ref{drs1}.

Combining the remark on the ellipsis ``on Monday'' and the DRS for the
dialog so far, we find that the right-hand argument of the operator
{\it or} is $\alpha$[Saturday/Monday]\footnote{i.e.\ in $\alpha$ all
appearences of Saturday are substituted by Monday}.
So, we can denote a DRS (see Fig.\ \ref{drs2}) that describes the
semantics of $\epsilon$.

\begin{figure}[htb]
$$
\epsilon=\left[
\begin{array}{l}
X\,\alpha[\mbox{Saturday/Monday}]\\\hline
\alpha=\lambda\,f.
\left[
\begin{array}{l}
\mbox{s}\,\mbox{r}\,f\\\hline
\mbox{\sc flight}(f)\\
\mbox{\sc Rome}(\mbox{r})\\
\mbox{\sc to}(f,\mbox{r})\\
\mbox{\sc Saturday}(\mbox{s})\\
\mbox{\sc on}(f,\mbox{s})
\end{array}
\right]\\
X=?\\
X\,\mbox{\bf or}\,\alpha[\mbox{Saturday/Monday}]
\end{array}
\right]
$$
\caption{\label{drs2}DRS for $\epsilon$}
\end{figure}

Obviously, as can be seen from the DRS for $\epsilon$, the problem of
finding an appropriate discourse referent for $X$ can be solved by
anaphora
resolution in the {\it describing situation}. The only antecedent in
the {\it describing situation} compatible with
$\alpha[\mbox{Saturday/Monday}]$ is $\alpha$. So, $\alpha$ has been
centered by $\epsilon$.

\begin{figure}
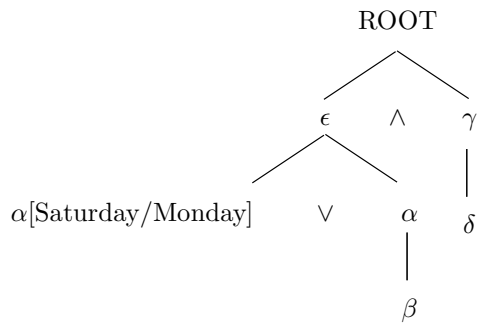

\centerline{\input or-subst.pstex_t}
\caption{\label{sub-by-or}Dialog Tree after Applying
$\alpha\,\mbox{\bf or}\,\alpha[\mbox{Saturday/Monday}]$}
\end{figure}

The impact of $\epsilon$ on the dialog structure is determined
essentially by how {\it or} modifies the dialog segmentation: the
segment of $\alpha$ is substituted by an OR subtree representing the
two arguments of {\it or} (see Fig.\ \ref{sub-by-or}). For the
utterances to follow, $\epsilon$ is the center, and dialog processing
works as usual.

\subsection{Performatives and Modal Verbs}
\label{performatives}
Performatives and modal verbs state assertions not about the {\it
described}, but the {\it describing situation}; more precisely, they
express assertions about mental states and speech acts as in ``I must
leave you now'', ``On what day do you want to depart?'', ``I suggest
not to pay at all for this bad film.''

As such utterances do not talk about the {\it described situation},
they cannot be processed as if they did. Consequently, speech act
recognition does not apply as normally in this case. For that reason,
all rules above for inferring speech acts are defeasible. Therefore we
can devise ``special'' rules for performatives and modal verbs that
``override'' defeasible inferences based solely on syntactic and
prosodic criteria if an inspection of the content of the utterance
provides evidence against these ``default conclusions''.
To do this, we classify performatives and modal verbs according to
what mental state they operate on or what speech act they express.

In the utterance ``I want to fly to Rome on Saturday.'', {\it want}
expresses implicitly the query ``Is there a flight to Rome on
Saturday.''. It is clear immediately that the utterance is
actually not an {\sf inform} act that poses no obligations to respond
cooperatively on the hearer. So desires and queries are defeasibly
connected by:
\begin{eqnarray}
\des{I}{\action{I}{K}}&\rightarrow&\ast(\query I{K},\query I{K})
\end{eqnarray}
$\des I{\action I{K}}$ overrides inform in the following way:
\begin{eqnarray}
\label{notinform}
\des I{\action I{K}}&\rightarrow & \neg \inform I{K}
\end{eqnarray}
When the dialog manager starts processing the utterance above, it
infers the following:
\begin{itemize}
\setlength{\parsep}{0pt}
\setlength{\itemsep}{0pt}
\item $\inform I{K}$ with {\it justification context} $\inform I{K}$
\item $\query I{K}$ with {\it justification context} $\query I{K}$
\end{itemize}
As the semantics of ``I want to'' belongs to the class $\des I{\action
I{K}}$, we can use this fact (that has been derived from the semantic
representation of the utterance) to infer $\neg \inform I{K}$ by
(\ref{notinform}).
As a result, the only valid inference is $\query I{K}$. By
cooperation, we can infer $\des R{\know I{K}}$. After that, we are in
exactly the same situation as if the speaker had asked directly for a
flight to Rome.

\section{Conclusions}
We have reported about work in progress on developing
a theoretical approach on dialog structure in order to build domain
independent, cooperative, and robust dialog managers. The theoretical
framework outlined in this paper has already been implemented partially
and
was tested for a small domain. In an evaluation, the
implementation has proven to perform well.
On the other hand, we have analyzed a large corpus of train-inquiry
dialogs collected with our EVAR (\cite{EckertNiemann:94}) system. An
important result is that the quality of speech recognition determines
how cooperative the EVAR system\footnote{EVAR is not based on the
dialog model sketched in this paper.} really is. To improve this
situation we work towards improving dialog theory as presented
in this paper by integrating
BDI-oriented, structural, and plan based approaches to dialog
understanding. Experience from analyzing dialogs that have not been
terminated successfully has shown that our approach is capable to
overcome the failures that caused the inacceptable terminations.

\section{Future Research}
Using the results presented in this paper as
a basis, we will continue our work on dialog structure by describing
precisely how a processing level for handling typically difficult
phenomena of spoken language like repairs etc.\ can be integrated into
our model. We intend to achieve this by incorporating Traum's model of
grounding (see \cite{Traum:94}) into our framework. This
mechanism will have to be expanded to work properly on word
hypothesis graphs that are the basic data structure of the common
ground. This allows
for a full exploitation of the results produced by the speech
recognizer. Furthermore, we want to integrate our implementation of a
chunk parser as a robust algorithmic approach to a theory of
incremental discourse processing as described e.g.\ in
\cite{Poesio:94}.
{%\twocolumn
%\raggedright
}
\end{document}